\documentclass{article}
\pdfoutput=1

     \usepackage[preprint]{neurips_2020}

\usepackage{wrapfig}
\usepackage{wrapfig}
\usepackage{graphicx}
\usepackage{multirow}
\usepackage[utf8]{inputenc} 
\usepackage[T1]{fontenc}    
\usepackage{hyperref}       
\usepackage{url}            
\usepackage{booktabs}       
\usepackage{amsfonts}       
\usepackage{nicefrac}       
\usepackage{microtype}      

\title{Stereo RGB and Deeper LIDAR Based Network for 3D Object Detection}

\author{%
  Qingdong He,\ Zhengning Wang\thanks{Corresponding author.}\  , Hao Zeng, Yijun Liu, Shuaicheng Liu, Bing Zeng\\
   University of Electronic Science and Technology of China \\
  Chengdu, China\\
  \texttt{\{heqingdong,haozeng,liuyijun\}@std.uestc.edu.cn}\\
  \texttt{\{zhengning.wang,liushuaicheng,eezeng\}@uestc.edu.cn} \\
}

\begin{document}

\maketitle

\begin{abstract}
  3D object detection has become an emerging task in autonomous driving scenarios. Previous works process 3D point clouds using either projection-based or voxel-based models. However, both approaches contain some drawbacks. The voxel-based methods lack semantic information, while the projection-based methods suffer from numerous spatial information loss when projected to different views. In this paper, we propose the Stereo RGB and Deeper LIDAR (SRDL) framework which can utilize semantic and spatial information simultaneously such that the performance of network for 3D object detection can be improved naturally. Specifically, the network generates candidate boxes from stereo pairs and combines different region-wise features using a deep fusion scheme. The stereo strategy offers more information for prediction compared with prior works. Then, several local and global feature extractors are stacked in the segmentation module to capture richer deep semantic geometric features from point clouds. After aligning the interior points with fused features, the proposed network refines the prediction in a more accurate manner and encodes the whole box in a novel compact method. The decent experimental results on the challenging KITTI detection benchmark demonstrate the effectiveness of utilizing both stereo images and point clouds for 3D object detection.
\end{abstract}

\section{Introduction}

Accurate and robust 3D object detection is crucial and indispensable for many real-world applications, such as autonomous driving~\cite{geiger2012we} and augmented reality (AR) ~\cite{Park:2008:MOT:1605298.1605357}. State-of-the-art methods can achieve a high average precision (AP) of 2D object detection ~\cite{ren2015faster,redmon2016you} and have achieved honorable results in the testing of public datasets such as KITTI ~\cite{geiger2012we} and COCO ~\cite{chen2015microsoft}. However, directly extends 2D detection methods to 3D is nontrivial due to the sparseness and irregularity of the point clouds. How to process the point clouds data with the semantic information from the RGB data remains a hot and challenging problem.

Currently, researchers have explored several methods to tackle these problems, which aim to obtain geometric information such as target position, size and posture in 3D space. Some works~\cite{chen2016monocular,mousavian20173d,xu2018multi,chen20173d,li2018stereo} make full use of the characteristics of RGB images to propose some networks. However, the key problem of the image-based methods is that the depth information cannot be directly obtained, which results in a large positioning error of the object in 3D space. Even stereo vision~\cite{li2019stereo} is very sensitive to factors such as illumination variations and occlusions, which lead to deviations in depth calculations.

Compared with image data, LIDAR point clouds data have accurate depth information and spatial features. At present, most of the state-of-the-art 3D object detection algorithms have focused on processing LIDAR point clouds through projection \cite{simon2019complexer,chen2017multi,ku2018joint,liang2018deep,yang2018hdnet,yang2018pixor} or voxelization \cite{li20173d,engelcke2017vote3deep,zhou2018voxelnet}. However, these works either suffer from the information loss during projectionand quantization \cite{liu2019point} or heavily depend on the performance of 2D object detectors\cite{qi2018frustum}. Recently, some works~\cite{shi2019pointrcnn,yang2019std,Point-GNN} propose to only operate on the point clouds to fulfill 3D object detection. But they achieve an inferior performance especially on cyclist and pedestrian due to the information loss from the image plane.

Different from aforementioned methods, we observed that stereo camera can provide large-scale perception from two views and LIDAR sensors can capture accurate 3D structures, while the combination of them could take advantage of their respective advantages while making up for their shortcomings. In other words, the left and right images can provide a more accurate receptive field while achieving comparable depth and position accuracy. Furthermore, we find that the most commonly used PointNet~\cite{qi2017pointnet,qi2017pointnet++} fails to capture local features information at variable scales and leads to the loss of local features because it only processes 3D points independently to maintain permutation invariance. In this way, it ignores the distance metric between the points. Although the latter SAWnet ~\cite{kaul2019sawnet} integrates the global features using shared Multi-Layer Perceptron (MLP) with the dynamic locality information from Dynamic Graph CNNs (DGCNNs) ~\cite{Wang:2019:DGC:3341165.3326362}, it is unable to focus on important features and suppress unnecessary ones in its residual connections ~\cite{he2016deep}.

\par Motivated by these observations, we present the Stereo RGB and Deeper LIDAR (SRDL) network for 3D object detection, which takes stereo RGB images and LIDAR point clouds as input and utilizes attention mechanism to achieve robust and accurate 3D detection. Specifically, the left and right views can generate proposals that do not completely overlap from different angles. They can mutually correct each other, and a more precise region can be generated during the fusion phase.
Considering that the fused proposals may overlap noisy points and excess space for the objects, a feature-oriented segmentation network in 3D point clouds is designed to strip out the object point clouds from the background. Given the segmented object points and cropped proposals, we propose to encode the bounding boxes by adding more constraints in a novel compact manner. This design benefits for removing more redundancy and locating the size of the objects more accurately while reducing the feature dimensions.

The main contributions of our work can be summarized as follows:
\begin{itemize}
\item To the best of our knowledge, we are the first to propose a novel framework that combines semantic information from stereo images and spatial information from raw point clouds for 3D object detection.
\item We propose residual attention learning mechanism to optimize the segmentation network, which can extract deeper geometric features of different levels from the original irregular 3D point clouds.
\item We propose a novel 3D bounding box encoding scheme that regresses the oriented 3D boxes in a more compact manner, ensuring higher 3D localization accuracy.
\item Our proposed SRDL network achieves comparable results with the state-of-the-art image-based and LIDAR-based methods on the challenging KITTI 3D detection dataset.
\end{itemize}

\section{Related work}
{\bf Image-based 3D object detection.} For processing the RGB images, there are two mainstreams, monocular-based and stereo-based methods. In terms of monocular-based methods, many researches ~\cite{ma2019accurate,chen2016monocular,mousavian20173d,brazil2019m3d} have contributed to share similar framework with 2D detection. Surprisingly, there are only a few works utilizing stereo vision for 3D object detection ~\cite{chen20173d, li2018stereo}. Typically, Li et al ~\cite{li2019stereo} propose the Stereo RCNN to detect and associate object in stereo images by both semantic properties and dense constraints of objects, extending Faster RCNN ~\cite{ren2015faster} for stereo inputs. However, none of the above approaches combines stereo images with point clouds properly to exploit both advantages and they fail to achieve superior performance because of the lack of accurate depth information.

{\bf LIDAR-based object detection.} Generally, there are two major ways to process the point clouds from 3D LIDAR sensors, voxelization and raw point clouds. The voxelization based methods ~\cite{liu2019tanet,zhou2018voxelnet,yan2018second,shi2019pv} usually take the voxel as input and apply either 2D convolution or 3D convolution to make prediction. VoxelNet ~\cite{zhou2018voxelnet} is one of the first methods to apply a PointNet-like network to learn low-level geometric feature with several stacked VFE layers in 3D voxelization space. However, the network structure is computationally inefficient as the shallow 3D CNN layers ~\cite{li20173d} are not enough to extract deeper 3D features. Even though SECOND ~\cite{yan2018second} applies sparse convolution to accelerate VoxelNet, it is still unable to overcome the 3D convolution bottleneck.
\par Besides, PointNet ~\cite{qi2017pointnet} and PonintNet++ ~\cite{qi2017pointnet++} are the two pioneers to directly operate on raw points to extract features without converting them to other formats. Based on PointNet as backbone network, some researchers have approached to infer 3D objects from point clouds ~\cite{shi2019pointrcnn,chen2019fast,lang2019pointpillars,shi2019part,yang2019std}. Very recently, Point-GNN\cite{Point-GNN} even propose graph neural network to  to detect objects from point clouds.

{\bf LIDAR and RGB image fusion based object detection.} The majority of the state-of-the-art 3D object detection methods adopt a LIDAR and mono-image fusion scheme to provide accurate 3D information, where they process raw LIDAR input in different representations. Many methods~\cite{chen2017multi,ku2018joint,liang2018deep,yang2018pixor,yang2018hdnet} project point clouds to bird’s view or front view and utilize 2D CNN to obtain more dense information for 3D box generation. However, these methods still have the limitation when detecting small objects such as pedestrians and cyclists and they do not deal with cases with multiple objects in depth direction. F-PointNet ~\cite{qi2018frustum} is the first method of utilizing mature 2D detectors and raw point cloud to predict 3D objects. And PointNet then is employed to process point cloud within every cropped image region to detect 3D objects. However, the mono-image can't extract more comprehensive features better than binocular and PointNet lacks the ability to capture local feature information in the metric space.
\section{Proposed method}
\label{sec:sim}
\begin{figure*}[ht]
\centering
  \includegraphics[height=4cm,width=12cm]{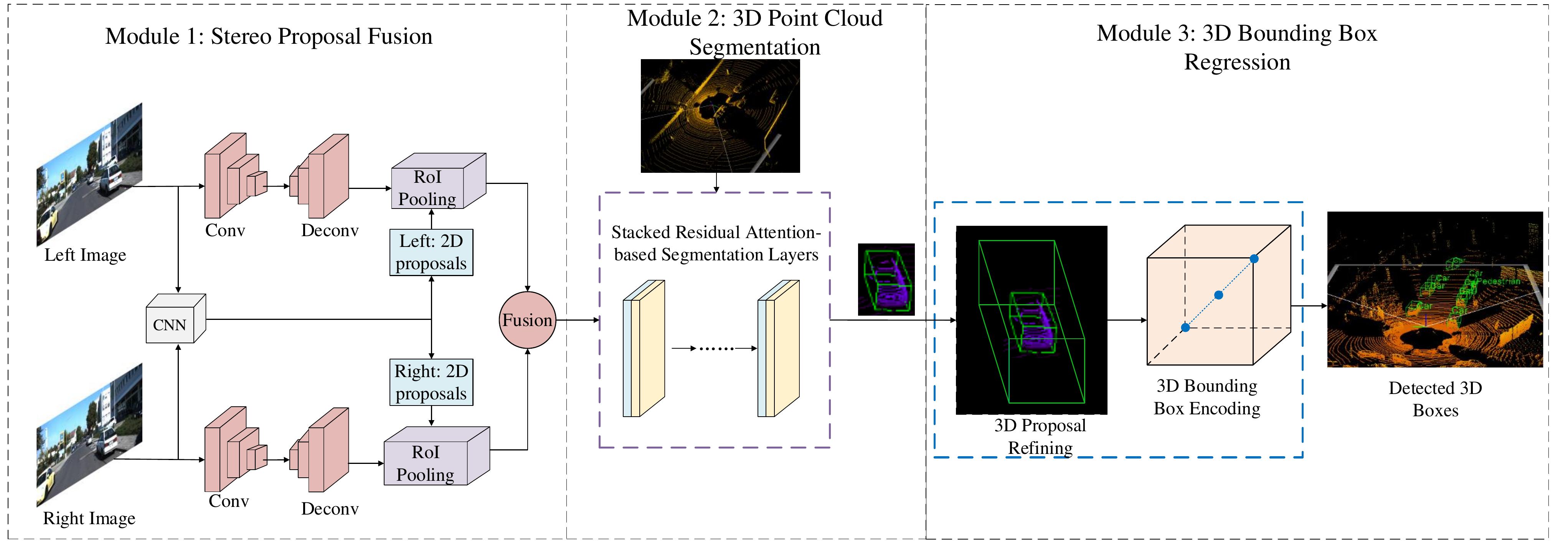}
  \caption{Architecture of the proposed SRDL network which contains three modules: (a) Stereo proposal fusion, which takes stereo RGB images as input and utilizes CNN to generate 2D proposals following RoIpooling in the two views respectively. (b) The 3D point clouds segmentation stacks several attention-based layers to separate the points of objects from backgrounds according to the projected proposals after the fusion operation. (c) After refining the boxes, the 3D bounding box regression module proposes to encode the bounding box in a more accurate scheme to get the final detection results.}
  \label{fig:1}
\end{figure*}
\begin{wrapfigure}{r}{6.5cm}
\centering
\includegraphics[width=0.4\textwidth]{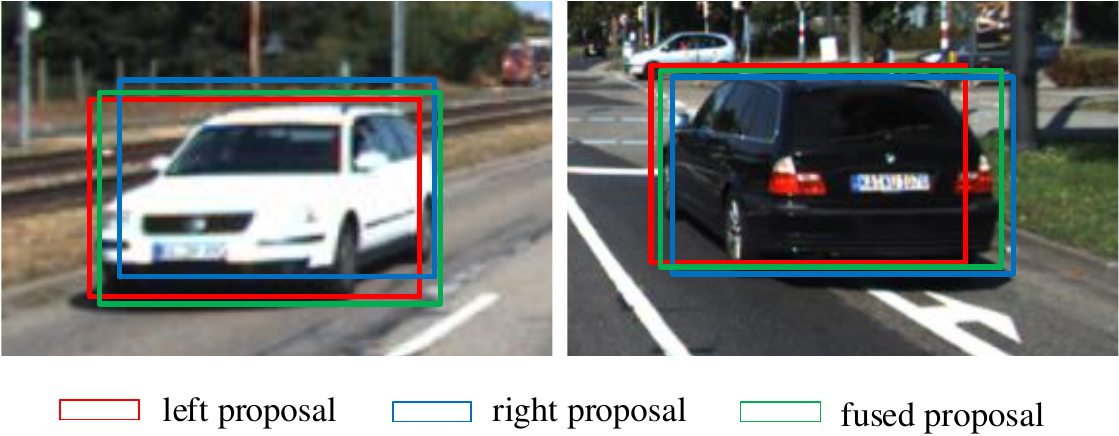}
\caption{ Proposals from left and right views. The proposals from the left and right views are not completely overlapped and the final fused proposal is more accurate than either of them.}
\label{fig:2}
\end{wrapfigure}
In this paper, we propose the Stereo RGB and Deeper LIDAR (SRDL) network for 3D object detection, including three modules: stereo proposal fusion, 3D point clouds segmentation and 3D bounding box regression, as shown in Figure~\ref{fig:1}. In the flowing subsections, we will introduce these modules in detail.
\subsection{Stereo proposal fusion}
In our framework, we take the stereo images as input and leverage mature 2D object detector to generate 2D object proposals for the left and right image respectively. At the same time, we apply convolution-deconvolution in each view to acquire features in a higher resolution. Combining the 2D proposals, RoIpooling is employed for each view to obtain features at the same size. Finally, we fuse the two cropped features of the output in the RoIpooling via element-wise mean operation.

As Figure~\ref{fig:2} shows, the two outputs of the left and right branches are not completely overlapped. Instead, each of them generates different proposals from different views. With a known camera projection matrix to offer accurate depth information, each bounding box can be projected into 3D space to form a cross object area. Through the final element-wise fusion, the final proposal contains less space and point clouds which is more accurate than either of the initial ones by mutual supervision and correction.
\subsection{3D point cloud segmentation}
\label{sec:sim4}
As illustrated in Figure~\ref{fig:1}, the fused proposal is fed into the second module with the depth range to outline an appropriate location in 3D space. Given the 2D image region and its corresponding 3D locations, we design a 3D segmentation network to sperate 3D point clouds from background for further 3D coordinate regression.

\subsubsection{Architecture overview}
The input of the segmentation network architecture is fed into a transformer net that uses attention-based layers to regress a $4\times4$ transformation matrix, the elements of which are the learnt affine transformation values for point clouds alignment. The aligned points are then fed into several stacked attention-based layers to generate a permutation invariant embedding of the points. Among them, the residual attention module serves as a linking bridge between two adjacent layers to transfer information. After that, the outputs of all the previous 1024-D attention-based layers are concatenated together and the max pooling is used to get the final global information aggregation for the point clouds. The information is then fed into a MLP layer to predict a $N\times p$ score matrix and make a point-wise prediction. More details of the specific architecture are described in the supplementary.

\subsubsection{Attention-based layer}
\label{sec:sim2}
\begin{wrapfigure}{r}{6cm}
\centering
\includegraphics[height=7.3cm,width=4.5cm]{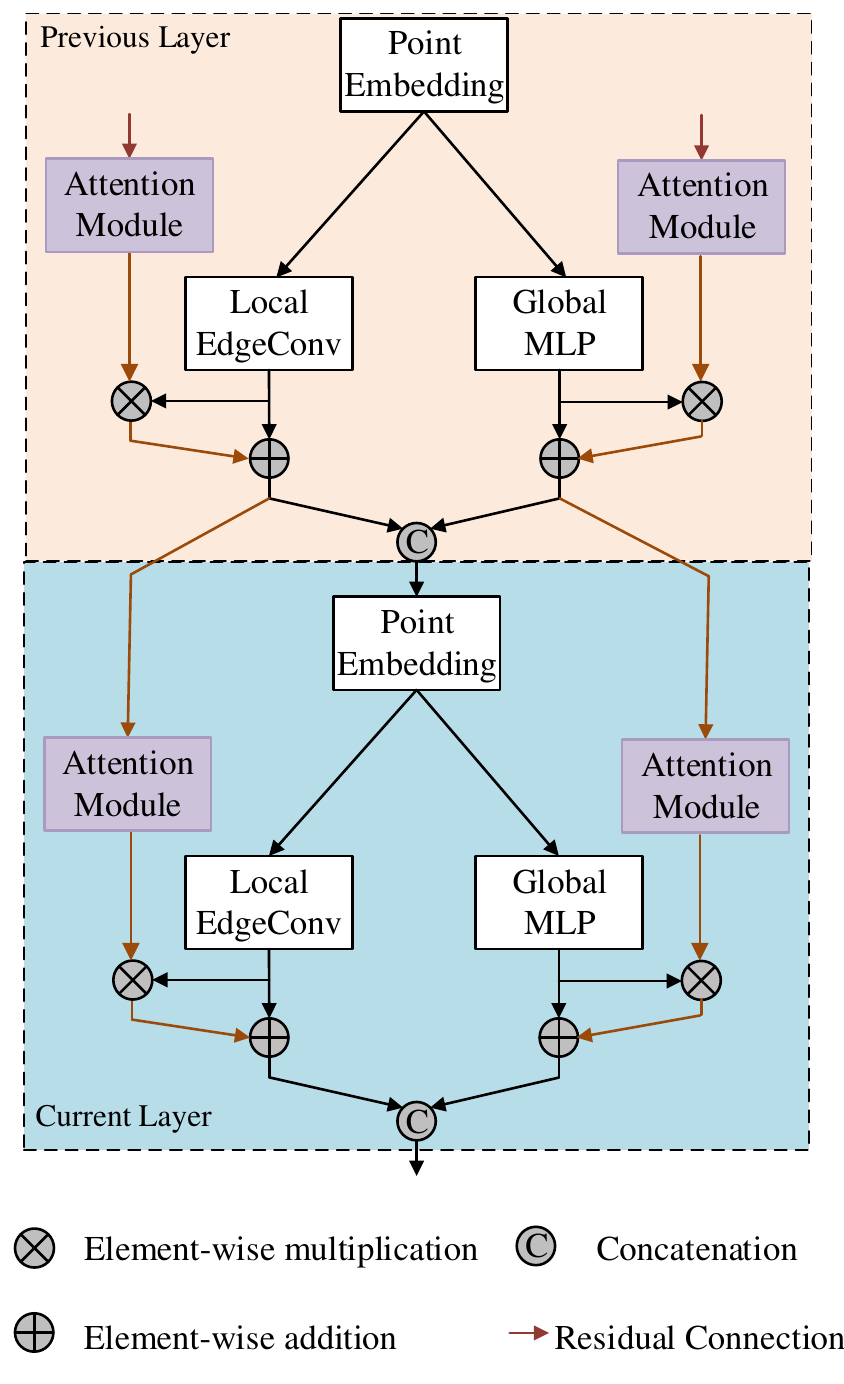}
\caption{ The illustration of the information propagates between two adjacent layers. There are residual-attention connections to transfer local and global information from previous layer to current layer individually.}
\label{fig:3}
\end{wrapfigure}
The architecture of the attention-based layers is shown in Figure~\ref{fig:3}, in which the current layer is considered as an intermediate layer and the features are not only transmitted by the mainstream information flow but the residual attention linking from previous layer. The point embedding from the current is input into two parallel layers, local EdgeConv layer and global MLP layer respectively. The local EdgeConv layer constructs a dynamic graph and incorporates $k$ nearest local neighborhood information. The global MLP layer operates on each point independently and subsequently applies a symmetric function to accumulate features. The outputs of these two layers are connected to the outputs of the same branch of the previous layer in an element-wise manner before concatenating together. The two layers also individually transfer information to the next embedding layer using residual attention connections.
\par Specifically, consider $n$ points in a $D$ dimensional embedding point clouds set $P=\{p_1,p_2,...,p_n\}$, where $D$ can be set to 3 simply, which means each point $x_n$ contains three coordinates $(x_i,y_i,z_i)$. These points are processed in parallel by the local EdgeConv layer and global MLP layer in each attention-based layer. In the branch of the local EdgeConv layer, let $h(k)$ denote the input in terms of the $k$ nearest neighbours in the dynamic graph and $e_i(i=1,2)$ denote the edgeconv operation to evaluate a point's dependency on its $k$ nearest neighbours. The extracted edge features are fed into the batch normalization and ReLU computation. The output can be represented as:
\begin{equation}
\label{eqn:01}
E_1=MLP(h(k))=MLP(e_1(h(k)))=\sigma(W_1(h(k))).
\end{equation}
After applying another edgeconv-BN layer, the output can be represented as:
\begin{equation}
\label{eqn:01}
E_2 =MLP(E_1)=MLP(e_2(E_1))=W_2(E_1).
\end{equation}
Where $\sigma$ denotes the ReLU function and $W_1$, $W_2$ are the weights of the two MLPs.
After the maxpooling over the output, the attention map from the residual module is added to the output $E_2$ in a point-wise manner which can be written as:
\begin{equation}
\label{eqn:01}
L=(1+R_1) \otimes E_2.
\end{equation}
Similarly, in the global MLP layer, $f(t)$ denotes the transformation on the input points by the shared weighted MLP denoted as $s$. The output of the first shared MLP layer is:
\begin{equation}
\label{eqn:01}
M_1=MLP(f(t))=\sigma(W_1(s(k))).
\end{equation}
Applying another MLP-BN layer, the output is:
\begin{equation}
\label{eqn:01}
M_2=MLP(M_1)=W_2(s(M_1)).
\end{equation}
This output is connected to attention-aware feature in the residual attention module in the same way:
\begin{equation}
\label{eqn:01}
G=(1+R_2)\otimes M_2.
\end{equation}
Where $\otimes $ denotes element-wise multiplication and $R_i(i\in\{1,2\})$ change between 0 and 1 as the reaction to different features. Different from the original ResNet, the output of our residual attention module $R_i(i\in\{1,2\})$ works as the feature filter to weaken the noisy features and amplify the good features.
Note that the outputs $L, G$ from the two branches have the same dimension and are transferred to the next layer as well. Finally, they are concatenated together and the embedding points are fed into the next layer as the input.

\subsubsection{Attention module}
\begin{wrapfigure}{r}{5cm}
\centering
\includegraphics[height=6.1cm,width=3.8cm]{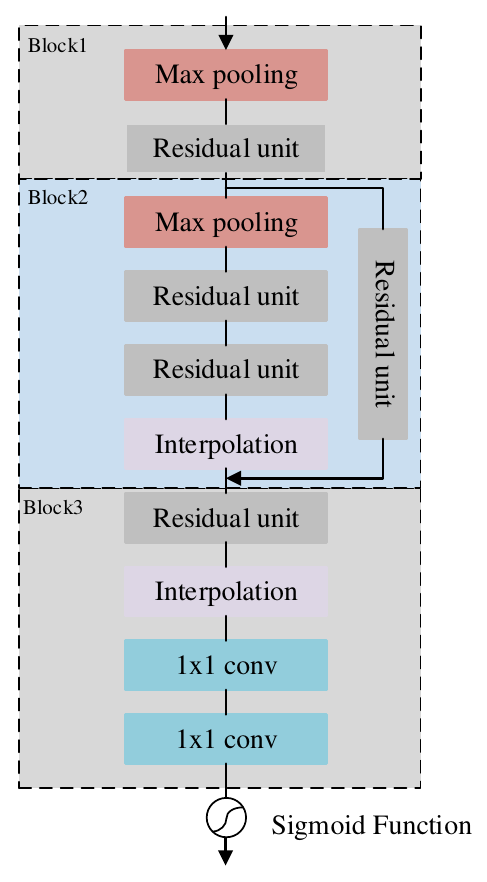}
\caption{ Architecture of attention module which mainly consists of residual units in a bottom-up and top-down manner.}
\label{fig:4}
\end{wrapfigure}
Attention module not only attempts to emphasize meaningful features but also enhances different representations of objects at certain locations. We design our attention module as a bottom-up and a top-down structure, as shown in Figure~\ref{fig:4}. The bottom-up operation aims to collect global information and the top-down operation combines the global information with the original feature maps. We use the residual unit in ~\cite{he2016identity} as our basic unit in attention module. The attention module contains three blocks. In block1, the max pooling and a residual unit are performed to enlarge the receptive field. After getting the lowest resolution, a symmetrical top-down architecture is designed to infer each pixel to get dense features in block2. Besides, we append skip connections between bottom-up and top-down feature maps to capture features at different scales. In block3, a bilinear interpolation is inserted after a residual unit to up-sample the output. Finally, we use the sigmoid function to normalize the output after two consecutive $1\times1$ convolution layers to balance the dimensions.
\subsection{3D bounding box regression}
Given the segmented object points, this part regresses the final 3D bounding box by a more accurate bounding box encoding scheme after the proposal refining.
\subsubsection{3D proposal refining}
\label{sec:sim1}
After segmentation operation on the point cloud, we can sperate the object points from the background and acquire the points inside the bounding box in the certain location of the first module. However, the combination of the predefined proposals from the first module and the segmentation network for the points only gets a relatively rough box. Therefore, we propose to pool 3D points and their corresponding features to rescale the proposal. For each 3D box proposal, $b_i=(x_i,y_i,z_i,w_i,h_i,l_i,\theta_i)$, we define a new 3D box by adding a constant $\xi$ to $w_i,h_i,l_i$ respectively to resize the box. For each point, a validation test is performed to decide whether it is inside the resized box or not. If it is true, the point and its features will be kept for refining the box proposal. Further ablation study will illustrate the effectiveness of this operation in improving performance.
\subsubsection{3D bounding box encoding}
\label{sec:sim3}
\begin{wrapfigure}{r}{7cm}
\centering
\includegraphics[width=0.5\textwidth]{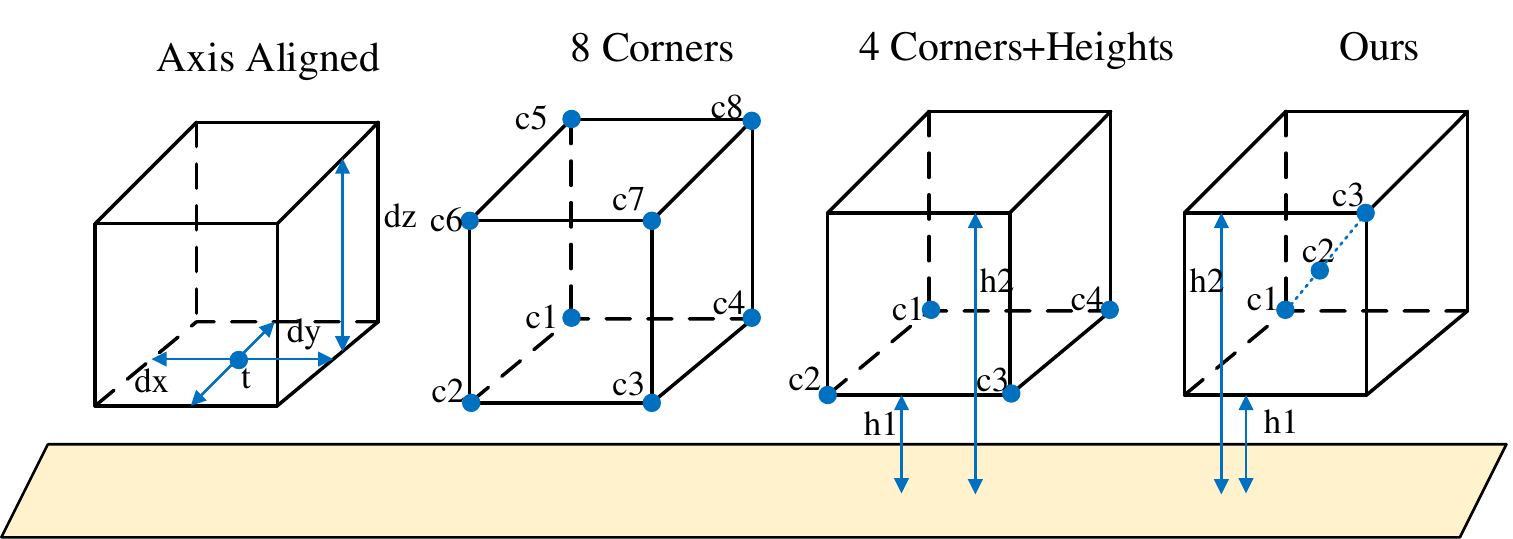}
\caption{ Comparison between different methods for encoding bounding box. We propose to encode the bounding box with three points (two corners + one center) and two heights to reduce redundancy and keep physical connections.}
\label{fig:5}
\end{wrapfigure}
To determine the orientation of a 3D bounding box, we keep consistent with the AVOD which computes $(\cos\theta,\sin\theta)$ to solve the problem of angles wrapping. As for the box encoding, there are several different methods to encode the bounding box as shown in Figure~\ref{fig:5}. The axis aligned is first proposed in ~\cite{song2016deep} which encodes the box with centers and sizes. While in MV3D~\cite{chen2017multi}, Chen et al. claim that 8 corners box encoding works better than axis aligned. And in AVOD~\cite{ku2018joint}, Jason Ku et al. attempt to replace 8 corners with 4 corners and 2 heights to encode box efficiently. However, 8 corners need a 24-D vector to normalize the diagonal length of the proposal box and neglect the physical constraints. The 4 corners + heights encoding method does not take the physical connections between the 4 corners within a plane into account. To reduce more redundancy and keep physical connections, we propose to encode the bounding box with three points (two corners + one center) and two heights representing the offsets from the proposal box to the ground plane. The three points are on the diagonal of the cube, where $c_2$ is the center point of the cube. Therefore, the regression targets is $\{\Delta x_i,\Delta y_i,\Delta z_i,\Delta h_j;i=1,2,3;j=1,2\}$. Despite that our 11-D representation vector is slightly larger than the 10 dimensional one, we not only use fewer points but encode the bounding box compactly in the flowing constrains between these parameters.
When regressing $w,l$, we should consider:
\begin{enumerate}
\item $(\Delta x_{c1},\Delta x_{c3})\rightarrow w$, $(|\Delta x_{c2}-\Delta x_{c1}|,|\Delta x_{c3}-\Delta x_{c2}|)\rightarrow w$.
\item $(\Delta y_{c1},\Delta y_{c3})\rightarrow l$, $(|\Delta y_{c2}-\Delta y_{c1}|,|\Delta y_{c3}-\Delta y_{c2}|)\rightarrow l$.
\end{enumerate}
When regressing $h$, we should ensure that the following equations hold true:
\begin{enumerate}
\item $|\Delta z_{c3}-\Delta z_{c1}|=|\Delta h_2-\Delta h_1|\rightarrow h$.
\item $|\Delta z_{c2}-\Delta z_{c1}|=|\Delta z_{c3}-\Delta z_{c2}|=\frac{1}{2}|\Delta h_2-\Delta h_1|\rightarrow h$.
\end{enumerate}
where $\rightarrow$ denotes that there exists constrains in the regression process.
\subsection{Loss function}
We use a multi-task loss to train our network. Our total loss is composed of three main components from the three modules, the fused loss $L_{fuse}$, the segmentation loss $L_{seg}$ and the bounding box regression loss $L_{box}$ as:
\begin{equation}
\label{eqn:01}
L_{total}=\alpha L_{fuse}+\beta L_{seg}+\chi L_{box}
\\=\alpha(L_{cls\_1}+L_{reg\_1})+\beta L_{seg}+\chi(L_{cls\_2}+L_{reg\_2}).
\end{equation}
where weighting parameters $\alpha$, $\beta$ and $\chi$ are used to balance the relative importance of different parts, and their values are set to 1, 4 and 2 respectively. $L_{cls\_1}$ and $L_{cls\_2}$ are the object classification loss. $L_{reg\_1}$ and $L_{reg\_2}$ are box regression loss.
In our practice, we apply binary cross entropy for all classification loss. As for regression, we employ Smooth L1 loss for all bounding box and orientation vector regression. For segmentation, we use the focal loss~\cite{lin2017focal} to handle the imbalance problem. More details of the specific losses are described in the supplementary.
\section{Experiments}
We evaluate our method on the 3D detection benchmark and the bird's eye view detection benchmark of the KITTI test server~\cite{geiger2012we}. For evaluation, we use average precision (AP) metric to compare with different methods and use the official 3D IoU evaluation metrics of 0.7, 0.5, and 0.5 respectively for the categories of car, cyclist, and pedestrian. In this section, we will introduce the experimental results. More description about datasets, implementation and training details are specified in the supplementary.
\subsection{Comparing with state-of-the-art methods }

\begin{wraptable}{r}{8.5cm}
\setlength{\abovecaptionskip}{-5pt}
\setlength{\belowcaptionskip}{-0pt}
	\tiny
\caption{Performance comparison on KITTI 3D object detection for car, pedestrian and cyclists.The evaluation metrics is the average precision (AP) on the official test set.}
\label{tab:one}
\begin{center}
\setlength{\tabcolsep}{0.2mm}
\begin{tabular}{ccccccccccc} \toprule
            \multirow{2}{*}{Method} & \multirow{2}{*}{Modality}  & \multicolumn{3}{c}{$AP_{car}(\%)$} & \multicolumn{3}{c}{$AP_{pedestrian}(\%)$} & \multicolumn{3}{c}{$AP_{cyclist}(\%)$} \\
            \cmidrule(r){3-5} \cmidrule(r){6-8} \cmidrule(r){9-11}
                  &    & Easy & Moderate & Hard & Easy & Moderate & Hard & Easy & Moderate & Hard\\ \hline
M3D-RPN\cite{brazil2019m3d} &    Mono    & 15.52 & 11.44  & 9.62 & -  & - & -  & - & - & -\\
 CE3R\cite{ma2019accurate}&    Mono    & 21.48 & 16.08  & 15.26 & -  & - & -  & - & - & -\\\hline
Stereo RCNN\cite{li2019stereo} &   Stereo   & 49.23 & 34.05  & 28.39 & -  & - & -  & - & - & -\\\hline
MV3D\cite{chen2017multi} & Mono+Lidar & 71.09 & 62.35  & 55.12 & -  & - & -  & - & - & -\\
F-Pointnet\cite{qi2018frustum} & Mono+Lidar & 81.20 & 70.39  & 62.19 & 51.21 & 44.89 & 40.23  & 71.96 & 56.77 & 50.39\\
AVOD-FPN\cite{ku2018joint} & Mono+Lidar & 81.94 & 71.88  & 66.38 & 50.80 & 42.81 & 40.88  & 64.00 & 52.18 & 46.61\\
F-ConvNet\cite{wang2019frustum} & Mono+Lidar & 85.88 &76.51 & 68.08 & 52.37 & 45.61 & 41.49 & {\bf79.58} & 64.68 & 57.03\\
MMF\cite{liang2019multi} & Mono+Lidar & 86.81 & 76.75  & 68.41 & -  & - & - & - & - & -\\\hline
Voxelnet\cite{zhou2018voxelnet} &   Lidar    & 77.47 & 65.11  & 57.73 & 39.48 & 33.69 & 31.51  & 61.22 & 48.36 & 44.37\\
SECOND\cite{yan2018second} &   Lidar    & 83.13 & 73.66  & 66.20 & 51.07 & 42.56 & 37.29  & 70.51 & 53.85 & 46.90\\
PointPillars\cite{lang2019pointpillars} &   Lidar    & 79.05 & 74.99  & 68.30 & 52.08 & 43.43 & 41.49 & 75.78 & 59.07 & 52.92\\
PointRCNN\cite{shi2019pointrcnn} &   Lidar    & 85.94 & 75.76  & 68.32 & 49.43 & 41.78 & 38.63 & 73.93 & 59.60 & 53.59\\
STD\cite{yang2019std} &   Lidar    & 86.61 & 77.63  & {\bf76.06} & 53.08 & 44.24 & 41.97 & 78.89 & 62.53 & 55.77\\
Point-GNN\cite{Point-GNN} & Lidar & 88.33 & 79.47 & 72.29 & 51.92 & 43.77 & 40.14 & 78.60 & 63.48 & 57.08\\\hline
      SRDL(ours) &Stereo+Lidar& {\bf89.27} & {\bf79.95}  & 73.79 & {\bf53.44}  & {\bf45.91} & {\bf42.61} & 78.68 & {\bf64.88} & {\bf57.74}\\
            \bottomrule
        \end{tabular}
\end{center}
\end{wraptable}
For the 3D object detection and the bird’s view detection test benchmark as shown in Table~\ref{tab:one} and Table~\ref{tab:two}, our proposed method achieves decent results compared with other state-of-the-art methods for all categories on three difficulty levels. For car, our method achieves better or comparable results than most of the methods. For the pedestrian and cyclist, SRDL gets large increases than the mono+Lidar methods due to combining stereo images, especially on the moderate and hard set. For the most important car category, we also report the performance of our method on KITTI val split and the results are shown in the supplementary.
\begin{wraptable}{r}{8.5cm}
\setlength{\abovecaptionskip}{-35pt}
\setlength{\belowcaptionskip}{-5pt}
	\tiny
\caption{Performance comparison on KITTI bird's eye view detection for car, pedestrian and cyclists. The evaluation metrics is the average precision (AP) on the official test set.}
\label{tab:two}
\begin{center}
\setlength{\tabcolsep}{0.4mm}
\begin{tabular}{ccccccccccc} \toprule
            \multirow{2}{*}{Method} & \multirow{2}{*}{Modality}  & \multicolumn{3}{c}{$AP_{car}(\%)$} & \multicolumn{3}{c}{$AP_{pedestrian}(\%)$} & \multicolumn{3}{c}{$AP_{cyclist}(\%)$} \\
            \cmidrule(r){3-5} \cmidrule(r){6-8} \cmidrule(r){9-11}
                  &    & Easy & Moderate & Hard & Easy & Moderate & Hard & Easy & Moderate & Hard\\ \hline
M3D-RPN\cite{brazil2019m3d} &    Mono    & 21.29 & 15.23  & 13.16 & -  & - & -  & - & - & -\\\hline
Stereo RCNN\cite{li2019stereo} &   Stereo   & 61.27 & 43.87  & 36.44 & -  & - & -  & - & - & -\\\hline
MV3D\cite{chen2017multi} & Mono+Lidar & 86.02 & 76.90  & 68.49 & -  & - & -  & - & - & -\\
F-Pointnet\cite{qi2018frustum} & Mono+Lidar & 88.70 & 84.00  & 75.33 & 58.09  & 50.22 & 47.20  & 75.38 & 61.96 & 54.68\\
AVOD-FPN\cite{ku2018joint} & Mono+Lidar & 88.53 & 83.79  & 77.90 & 58.75  & 51.05 & 47.54  & 68.09 & 57.48 & 50.77\\
F-ConvNet\cite{wang2019frustum} & Mono+Lidar & 89.69 & 83.08 & 74.56 & 58.90 & 50.48 & 46.72 & 82.59 & 68.62 & 60.62\\
MMF\cite{liang2019multi} & Mono+Lidar & 89.49 & 87.47  & 79.10 & -  & - & -  & - & - & -\\\hline
Voxelnet\cite{zhou2018voxelnet} &   Lidar    & 89.35 & 79.26  & 77.39 & 46.13  & 40.74 & 38.11  & 66.70 & 54.76 & 50.55\\
SECOND\cite{yan2018second} &   Lidar    & 88.07 & 79.37  & 77.95 & 55.10  & 46.27 & 44.76  & 73.67 & 56.04 & 48.78\\
PointPillars\cite{lang2019pointpillars} &   Lidar    & 88.35 & 86.10  & 79.83 & 58.66  & 50.23 & 47.19  & 79.14 & 62.25 & 56.00\\
PointRCNN\cite{shi2019pointrcnn} &   Lidar    & 89.47 & 85.58  & 79.10 & -  & - & -  & 81.52 & 66.77 & {\bf60.78}\\
STD\cite{yang2019std} &   Lidar    & 89.66 & 87.76  & {\bf86.89} & {\bf60.99}  & 51.39 & 45.89  & 81.04 & 65.32 & 57.85\\
Point-GNN\cite{Point-GNN} & Lidar & {\bf93.11} & 89.17 & 83.90 & 55.36 & 47.07 & 44.61 & 81.17 & 67.28 & 59.67\\\hline
      SRDL(ours) &Stereo+Lidar& 90.82 & {\bf89.74}  & 81.93 & 59.62  & {\bf51.46} & {\bf48.32}  & {\bf82.61} & {\bf69.11} & 60.37\\
            \bottomrule
        \end{tabular}
\end{center}
\end{wraptable}
\subsection{Qualitative results}
\begin{figure}[htbp]
\centering
  \includegraphics[width=14cm]{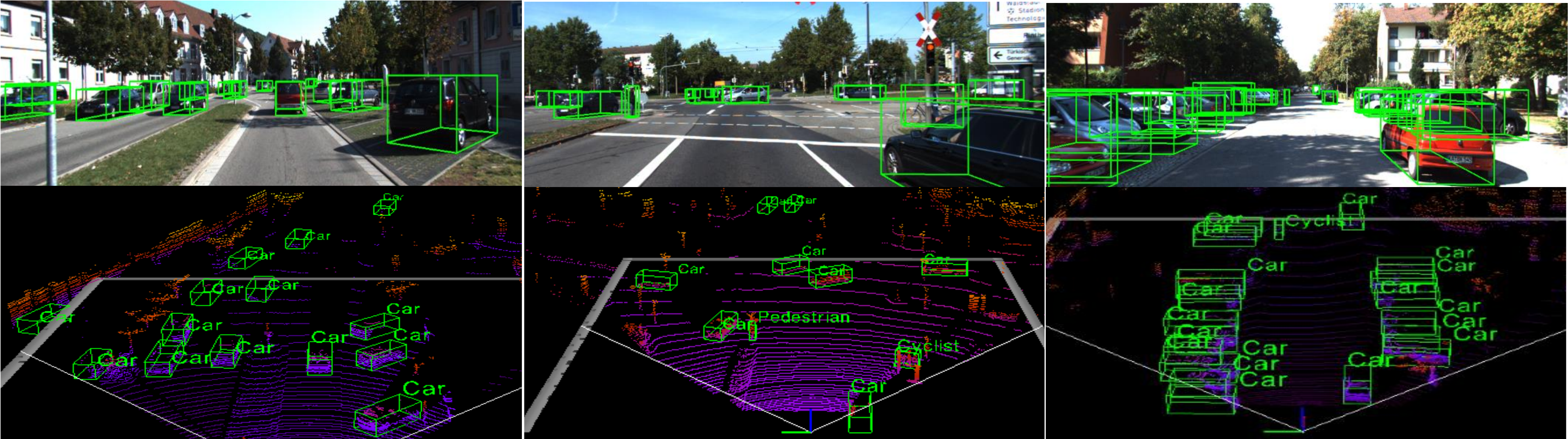}
  \caption{Qualitative 3D detection results of SRDL on the KITTI test set. The detected objects are shown with green 3D bounding boxes and the relative labels. The upper row in each image is the 3D object detection result projected onto the RGB image and the bottom is the result in the corresponding point clouds.}
  \label{fig:6}
\end{figure}

We present some qualitative results of our proposed SRDL network on the test split on KITTI dataset in Figure~\ref{fig:6}. From the figures we could see that our proposed network could estimate accurate 3D bounding boxes in different scenes. Surprisingly, we observe that our method can still achieve satisfactory detection results even with very sparse point clouds and severe occlusion.
\subsection{Ablation studies}
In this section, we change components and variants of our proposed SRDL by conducting extensive ablation studies on the validation  split of KITTI. We follow the convention and use the car class which contains the largest amount of training examples. The evaluation metric is the average precision (AP \%) on the val set.

{\bf Effect of Different Design Choice in the Whole Network.} We illustrate the importance of different components of our network by removing one part and keeping all the others unchanged, as shown in Table~\ref{tab:3}. Without the stereo images as input (the missing of “stereo” stands for mono image), the performance of SRDL drops dramatically which shows that the stereo images could provide rich feature information to locate the object. Similarly, AP decreases significantly by 11.65\%, 12.27\%, 16.33\% respectively for easy, moderate and hard which confirms the indispensability of the 3D bounding box encoding. And the performance degradation caused by the absence of either local or global convolution for segmentation proves that only the combination of them can produce the best results.
\begin{table}
\begin{minipage}[t]{0.35\linewidth}
\tiny
\caption{Performance of removing different part of our network. $\times$ denotes removing and $\checkmark$ denotes retaining.}
\label{tab:3}
\begin{center}
\setlength{\tabcolsep}{0.7mm}{
\begin{tabular}{ccccccc} \toprule
          Stereo &     Local    &     Global   &    Encoding   & Easy & Moderate & Hard \\\hline
        $\times$ & $\checkmark$ & $\checkmark$ & $\checkmark$  & 83.64 & 73.59 & 67.48 \\
    $\checkmark$ &  $\times$    & $\checkmark$ & $\checkmark$  & 86.77 & 72.32 & 71.65\\
    $\checkmark$ & $\checkmark$ & $  \times$   & $\checkmark$  & 88.46 & 76.71 & 75.69  \\
    $\checkmark$ & $\checkmark$ & $\checkmark$ & $\times$      & 78.63 & 67.55 & 62.32 \\
    $\checkmark$ & $\checkmark$ & $\checkmark$ & $\checkmark$  & {\bf90.28} & {\bf79.82} & {\bf78.65} \\
            \bottomrule
        \end{tabular}}
\end{center}
\end{minipage}%
\hfill
\begin{minipage}[t]{0.28\linewidth}
\tiny
\caption{Performance comparison of different fusion method with attention mechanism.}
\label{tab:4}
\begin{center}
\setlength{\tabcolsep}{0.6mm}{
\begin{tabular}{cccc} \toprule
            Fusion Method   & Easy & Moderate & Hard \\\hline
                 Global     & 83.85 & 71.08 & 65.73 \\
                  Local     & 83.89 & 71.27 & 66.54 \\
            Global+Local    & 84.51 & 71.33 & 68.79  \\
          Global+Attention  & 86.77 & 72.32 & 71.65 \\
           Local+Attention  & 88.46 & 76.71 & 75.69\\
    Global+Local+Attention  & {\bf90.28} & {\bf79.82} & {\bf78.65} \\
            \bottomrule
        \end{tabular}}
\end{center}
\end{minipage}
\hfill
\begin{minipage}[t]{0.25\linewidth}
\tiny
\caption{Performance comparison on different bounding box encoding methods.}
\label{tab:5}
\begin{center}
\setlength{\tabcolsep}{0.5mm}{
\begin{tabular}{cccc} \toprule
          Encoding Method   & Easy & Moderate & Hard \\\hline
                 Axis       & 79.13 & 68.42 & 65.36\\
              8 Corners     & 83.09 & 74.51 & 70.82 \\
       4 Corners+2 Heights  & 89.61 & 78.37 & 77.64  \\
        3 Points+2 Heights  & {\bf90.28} & {\bf79.82} & {\bf78.65} \\
            \bottomrule
        \end{tabular}}
\end{center}
\end{minipage}
\end{table}

{\bf Effect of Attention Module.} In order to show the importance of the attention module in Section~\ref{sec:sim2}, we add attention module to the three different designs. As shown in Table~\ref{tab:4}, with attention module to transfer features between connected layers, the fused models override the original ones by 1.24\%, 5.44\%, 8.49\% respectively in the moderate difficulty. And our final fusion method with attention mechanism outperforms the alternative by 5.77\%, 8.49\%, 9.86\% for the three difficulties.
\begin{wraptable}{r}{5.5cm}
\setlength{\abovecaptionskip}{-5pt}
\setlength{\belowcaptionskip}{-5pt}
\tiny
\caption{Performance of adopting different size of $\xi$ for 3D box refining. The "-" half denotes shrinking and the other half denotes enlarging. 0m is the original size.}
\label{tab:6}
\begin{center}
\setlength{\tabcolsep}{1mm}{
\begin{tabular}{cccc} \toprule
  refining size($\xi$)   & Easy & Moderate & Hard \\\hline
            1.5m    &  72.62 & 69.86 & 68.57 \\
            1.0m    &  79.43 & 70.53 & 70.82\\
            0.8m    &  84.59 & 72.58 & 71.94  \\
            0.5m    &  87.26 & 76.25 & 74.85 \\
             0m     &  89.47 & 78.84 & 77.62 \\
           -0.5m    & {\bf90.28} & {\bf79.82} & {\bf78.65} \\
           -0.8m    &  89.12 & 78.76 & 77.38 \\
           -1.0m    &  87.69 & 77.17 & 75.93 \\
           -1.5m    &  84.81 & 73.91 & 73.11 \\
            \bottomrule
        \end{tabular}}
\end{center}
\end{wraptable}
{\bf Effect of Different Refining Size $\xi$.} In Section~\ref{sec:sim1}, we propose to refine the proposals by adding a constant $\xi$ to the size of the box. Table~\ref{tab:6} shows the results with different size. $\xi$=-0.5m proves to perform best in our network which denotes that we should shrink the original box by 0.5m. Note that when we enlarge the size of the box, especially over 1m, the value of AP drops sharply. This indicates that the original box already contains redundant space, and continuing to enlarge the box will only include more unrelated areas. At the same time, too large $\xi$ to shrink the box also lead to bad performance since small region may also exclude relative areas.

{\bf Effect of Bounding Box Encoding Method.} As stated in Section~\ref{sec:sim3}, there are different bounding box encoding methods including the one we proposed. We use the four different methods to encode boxes in our proposed network. From Table~\ref{tab:5}, we note that although the 4 corners+2 heights method consumes a few dimensions but its performance is worse than our method. For one thing, the 4 corners+2 heights method does not take into account the coordinate relationship between the four corners so the number of points is redundant. For another thing, the constraint relationship between the coordinates of the corners and the heights cannot be established. Our method can establish four sets of constraint relationships to constrain the length, width, and height respectively.
\section{Conclusions}
In this paper, we have proposed a novel stereo RGB and deeper LIDAR (SRDL) network for 3D object detection in autonomous driving scenarios. Our method takes full advantage of the merits of stereo RGB images and point clouds to form an end-to-end framework. The combination of semantic information from stereo images and spatial information from point clouds contribute together to improve the performance. Extensive experiments on the challenging KITTI 3D detection benchmark demonstrate the efficiency of our method decently. In future research, we will optimize the inference speed, investigate more focus on integrating RGB and point-wise features, and different operations on the point clouds will be added to further improve our detection framework.
\section*{Broader Impact}
This article belongs to the application of a subtask under the study of autonomous driving. The accurate detection of vehicles and people on the road can greatly promote the development of autonomous driving. We know that our goal is to accurately detect 3D objects on the road to avoid various accidents. In this paper, stereo RGB images and point cloud data are used for joint detection. The dual data from the optical camera and radar camera jointly ensure this detection result, which is helpful for the further implementation of this application. For autonomous driving companies, they can find inspiration for further improving the safety of autonomous driving from the methods in this article, and this article also provides researchers with a new way to make full use of road environmental data. However, we have to admit that the method proposed in this article uses a variety of data, so the hardware requirements in the implementation are relatively high, which will bring some costs. In addition, once the data from a certain camera is missing as input, the algorithm in this paper will immediately fail.



\begin{ack}
This work is supported by a grant from the National Natural Science Foundation of China (No. 61872068, 61720106004), by a grant from Science \& Technology Department of Sichuan Province of China (No.2018GZ0071, 2019YFG0426), and by a grant from the Fundamental Research Funds for the Central Universities (No.2672018ZYGX2018J014).
\end{ack}


%
%
%
%
\small
\bibliographystyle{abbrv}
\bibliography{neurips_2020}

\begin{thebibliography}{10}

\bibitem{brazil2019m3d}
G.~Brazil and X.~Liu.
\newblock M3d-rpn: Monocular 3d region proposal network for object detection.
\newblock In {\em Proceedings of the IEEE International Conference on Computer
  Vision}, pages 9287--9296, 2019.

\bibitem{chen2015microsoft}
X.~Chen, H.~Fang, T.-Y. Lin, R.~Vedantam, S.~Gupta, P.~Doll{\'a}r, and C.~L.
  Zitnick.
\newblock Microsoft coco captions: Data collection and evaluation server.
\newblock {\em arXiv preprint arXiv:1504.00325}, 2015.

\bibitem{chen2016monocular}
X.~Chen, K.~Kundu, Z.~Zhang, H.~Ma, S.~Fidler, and R.~Urtasun.
\newblock Monocular 3d object detection for autonomous driving.
\newblock In {\em Proceedings of the IEEE Conference on Computer Vision and
  Pattern Recognition}, pages 2147--2156, 2016.

\bibitem{chen20173d}
X.~Chen, K.~Kundu, Y.~Zhu, H.~Ma, S.~Fidler, and R.~Urtasun.
\newblock 3d object proposals using stereo imagery for accurate object class
  detection.
\newblock {\em IEEE transactions on pattern analysis and machine intelligence},
  40(5):1259--1272, 2017.

\bibitem{chen2017multi}
X.~Chen, H.~Ma, J.~Wan, B.~Li, and T.~Xia.
\newblock Multi-view 3d object detection network for autonomous driving.
\newblock In {\em Proceedings of the IEEE Conference on Computer Vision and
  Pattern Recognition}, pages 1907--1915, 2017.

\bibitem{chen2019fast}
Y.~Chen, S.~Liu, X.~Shen, and J.~Jia.
\newblock Fast point r-cnn.
\newblock In {\em Proceedings of the IEEE International Conference on Computer
  Vision}, pages 9775--9784, 2019.

\bibitem{engelcke2017vote3deep}
M.~Engelcke, D.~Rao, D.~Z. Wang, C.~H. Tong, and I.~Posner.
\newblock Vote3deep: Fast object detection in 3d point clouds using efficient
  convolutional neural networks.
\newblock In {\em 2017 IEEE International Conference on Robotics and Automation
  (ICRA)}, pages 1355--1361. IEEE, 2017.

\bibitem{geiger2012we}
A.~Geiger, P.~Lenz, and R.~Urtasun.
\newblock Are we ready for autonomous driving? the kitti vision benchmark
  suite.
\newblock In {\em 2012 IEEE Conference on Computer Vision and Pattern
  Recognition}, pages 3354--3361. IEEE, 2012.

\bibitem{he2016deep}
K.~He, X.~Zhang, S.~Ren, and J.~Sun.
\newblock Deep residual learning for image recognition.
\newblock In {\em Proceedings of the IEEE conference on computer vision and
  pattern recognition}, pages 770--778, 2016.

\bibitem{he2016identity}
K.~He, X.~Zhang, S.~Ren, and J.~Sun.
\newblock Identity mappings in deep residual networks.
\newblock In {\em European conference on computer vision}, pages 630--645.
  Springer, 2016.

\bibitem{kaul2019sawnet}
C.~Kaul, N.~Pears, and S.~Manandhar.
\newblock Sawnet: A spatially aware deep neural network for 3d point cloud
  processing.
\newblock {\em arXiv preprint arXiv:1905.07650}, 2019.

\bibitem{ku2018joint}
J.~Ku, M.~Mozifian, J.~Lee, A.~Harakeh, and S.~L. Waslander.
\newblock Joint 3d proposal generation and object detection from view
  aggregation.
\newblock In {\em 2018 IEEE/RSJ International Conference on Intelligent Robots
  and Systems (IROS)}, pages 1--8. IEEE, 2018.

\bibitem{lang2019pointpillars}
A.~H. Lang, S.~Vora, H.~Caesar, L.~Zhou, J.~Yang, and O.~Beijbom.
\newblock Pointpillars: Fast encoders for object detection from point clouds.
\newblock In {\em Proceedings of the IEEE Conference on Computer Vision and
  Pattern Recognition}, pages 12697--12705, 2019.

\bibitem{li20173d}
B.~Li.
\newblock 3d fully convolutional network for vehicle detection in point cloud.
\newblock In {\em 2017 IEEE/RSJ International Conference on Intelligent Robots
  and Systems (IROS)}, pages 1513--1518. IEEE, 2017.

\bibitem{li2019stereo}
P.~Li, X.~Chen, and S.~Shen.
\newblock Stereo r-cnn based 3d object detection for autonomous driving.
\newblock In {\em Proceedings of the IEEE Conference on Computer Vision and
  Pattern Recognition}, pages 7644--7652, 2019.

\bibitem{li2018stereo}
P.~Li, T.~Qin, et~al.
\newblock Stereo vision-based semantic 3d object and ego-motion tracking for
  autonomous driving.
\newblock In {\em Proceedings of the European Conference on Computer Vision
  (ECCV)}, pages 646--661, 2018.

\bibitem{liang2019multi}
M.~Liang, B.~Yang, Y.~Chen, R.~Hu, and R.~Urtasun.
\newblock Multi-task multi-sensor fusion for 3d object detection.
\newblock In {\em Proceedings of the IEEE Conference on Computer Vision and
  Pattern Recognition}, pages 7345--7353, 2019.

\bibitem{liang2018deep}
M.~Liang, B.~Yang, S.~Wang, and R.~Urtasun.
\newblock Deep continuous fusion for multi-sensor 3d object detection.
\newblock In {\em Proceedings of the European Conference on Computer Vision
  (ECCV)}, pages 641--656, 2018.

\bibitem{lin2017focal}
T.-Y. Lin, P.~Goyal, R.~Girshick, K.~He, and P.~Doll{\'a}r.
\newblock Focal loss for dense object detection.
\newblock In {\em Proceedings of the IEEE international conference on computer
  vision}, pages 2980--2988, 2017.

\bibitem{liu2019point}
Z.~Liu, H.~Tang, Y.~Lin, and S.~Han.
\newblock Point-voxel cnn for efficient 3d deep learning.
\newblock In {\em Advances in Neural Information Processing Systems}, pages
  963--973, 2019.

\bibitem{liu2019tanet}
Z.~Liu, X.~Zhao, T.~Huang, R.~Hu, Y.~Zhou, and X.~Bai.
\newblock Tanet: Robust 3d object detection from point clouds with triple
  attention.
\newblock {\em AAAI}, 2020.

\bibitem{ma2019accurate}
X.~Ma, Z.~Wang, H.~Li, P.~Zhang, W.~Ouyang, and X.~Fan.
\newblock Accurate monocular 3d object detection via color-embedded 3d
  reconstruction for autonomous driving.
\newblock In {\em Proceedings of the IEEE International Conference on Computer
  Vision}, pages 6851--6860, 2019.

\bibitem{mousavian20173d}
A.~Mousavian, D.~Anguelov, J.~Flynn, and J.~Kosecka.
\newblock 3d bounding box estimation using deep learning and geometry.
\newblock In {\em Proceedings of the IEEE Conference on Computer Vision and
  Pattern Recognition}, pages 7074--7082, 2017.

\bibitem{Park:2008:MOT:1605298.1605357}
Y.~Park, V.~Lepetit, and W.~Woo.
\newblock Multiple 3d object tracking for augmented reality.
\newblock In {\em Proceedings of the 7th IEEE/ACM International Symposium on
  Mixed and Augmented Reality}, ISMAR '08, pages 117--120, Washington, DC, USA,
  2008. IEEE Computer Society.

\bibitem{qi2018frustum}
C.~R. Qi, W.~Liu, C.~Wu, H.~Su, and L.~J. Guibas.
\newblock Frustum pointnets for 3d object detection from rgb-d data.
\newblock In {\em Proceedings of the IEEE Conference on Computer Vision and
  Pattern Recognition}, pages 918--927, 2018.

\bibitem{qi2017pointnet}
C.~R. Qi, H.~Su, K.~Mo, and L.~J. Guibas.
\newblock Pointnet: Deep learning on point sets for 3d classification and
  segmentation.
\newblock In {\em Proceedings of the IEEE Conference on Computer Vision and
  Pattern Recognition}, pages 652--660, 2017.

\bibitem{qi2017pointnet++}
C.~R. Qi, L.~Yi, H.~Su, and L.~J. Guibas.
\newblock Pointnet++: Deep hierarchical feature learning on point sets in a
  metric space.
\newblock In {\em Advances in neural information processing systems}, pages
  5099--5108, 2017.

\bibitem{redmon2016you}
J.~Redmon, S.~Divvala, R.~Girshick, and A.~Farhadi.
\newblock You only look once: Unified, real-time object detection.
\newblock In {\em Proceedings of the IEEE conference on computer vision and
  pattern recognition}, pages 779--788, 2016.

\bibitem{ren2015faster}
S.~Ren, K.~He, R.~Girshick, and J.~Sun.
\newblock Faster r-cnn: Towards real-time object detection with region proposal
  networks.
\newblock In {\em Advances in neural information processing systems}, pages
  91--99, 2015.

\bibitem{shi2019pv}
S.~Shi, C.~Guo, L.~Jiang, Z.~Wang, J.~Shi, X.~Wang, and H.~Li.
\newblock Pv-rcnn: Point-voxel feature set abstraction for 3d object detection.
\newblock {\em arXiv preprint arXiv:1912.13192}, 2019.

\bibitem{shi2019pointrcnn}
S.~Shi, X.~Wang, and H.~Li.
\newblock Pointrcnn: 3d object proposal generation and detection from point
  cloud.
\newblock In {\em Proceedings of the IEEE Conference on Computer Vision and
  Pattern Recognition}, pages 770--779, 2019.

\bibitem{shi2019part}
S.~Shi, Z.~Wang, X.~Wang, and H.~Li.
\newblock Part-a\^{} 2 net: 3d part-aware and aggregation neural network for
  object detection from point cloud.
\newblock {\em arXiv preprint arXiv:1907.03670}, 2019.

\bibitem{Point-GNN}
W.~Shi and R.~R. Rajkumar.
\newblock Point-gnn: Graph neural network for 3d object detection in a point
  cloud.
\newblock In {\em The IEEE Conference on Computer Vision and Pattern
  Recognition (CVPR)}, June 2020.

\bibitem{simon2019complexer}
M.~Simon, K.~Amende, A.~Kraus, J.~Honer, T.~Samann, H.~Kaulbersch, S.~Milz, and
  H.~Michael~Gross.
\newblock Complexer-yolo: Real-time 3d object detection and tracking on
  semantic point clouds.
\newblock In {\em Proceedings of the IEEE Conference on Computer Vision and
  Pattern Recognition Workshops}, pages 0--0, 2019.

\bibitem{song2016deep}
S.~Song and J.~Xiao.
\newblock Deep sliding shapes for amodal 3d object detection in rgb-d images.
\newblock In {\em Proceedings of the IEEE Conference on Computer Vision and
  Pattern Recognition}, pages 808--816, 2016.

\bibitem{Wang:2019:DGC:3341165.3326362}
Y.~Wang, Y.~Sun, Z.~Liu, S.~E. Sarma, M.~M. Bronstein, and J.~M. Solomon.
\newblock Dynamic graph cnn for learning on point clouds.
\newblock {\em ACM Trans. Graph.}, 38(5):146:1--146:12, Oct. 2019.

\bibitem{wang2019frustum}
Z.~Wang and K.~Jia.
\newblock Frustum convnet: Sliding frustums to aggregate local point-wise
  features for amodal.
\newblock In {\em 2019 IEEE/RSJ International Conference on Intelligent Robots
  and Systems (IROS)}, pages 1742--1749. IEEE, 2019.

\bibitem{xu2018multi}
B.~Xu and Z.~Chen.
\newblock Multi-level fusion based 3d object detection from monocular images.
\newblock In {\em Proceedings of the IEEE Conference on Computer Vision and
  Pattern Recognition}, pages 2345--2353, 2018.

\bibitem{yan2018second}
Y.~Yan, Y.~Mao, and B.~Li.
\newblock Second: Sparsely embedded convolutional detection.
\newblock {\em Sensors}, 18(10):3337, 2018.

\bibitem{yang2018hdnet}
B.~Yang, M.~Liang, and R.~Urtasun.
\newblock Hdnet: Exploiting hd maps for 3d object detection.
\newblock In {\em Conference on Robot Learning}, pages 146--155, 2018.

\bibitem{yang2018pixor}
B.~Yang, W.~Luo, and R.~Urtasun.
\newblock Pixor: Real-time 3d object detection from point clouds.
\newblock In {\em Proceedings of the IEEE conference on Computer Vision and
  Pattern Recognition}, pages 7652--7660, 2018.

\bibitem{yang2019std}
Z.~Yang, Y.~Sun, S.~Liu, X.~Shen, and J.~Jia.
\newblock Std: Sparse-to-dense 3d object detector for point cloud.
\newblock In {\em Proceedings of the IEEE International Conference on Computer
  Vision}, pages 1951--1960, 2019.

\bibitem{zhou2018voxelnet}
Y.~Zhou and O.~Tuzel.
\newblock Voxelnet: End-to-end learning for point cloud based 3d object
  detection.
\newblock In {\em Proceedings of the IEEE Conference on Computer Vision and
  Pattern Recognition}, pages 4490--4499, 2018.

\end{thebibliography}

%
%
%

\end{document}